\documentclass[conference]{IEEEtran}
\IEEEoverridecommandlockouts
\usepackage{cite}
\usepackage{amsmath,amssymb,amsfonts}
\usepackage{algorithmic}
\usepackage{graphicx}
\usepackage{textcomp}
\usepackage{xcolor}

\usepackage{microtype}
\usepackage{booktabs}
\usepackage{subfigure}
\usepackage{accents}
\usepackage{listings}
\usepackage{stfloats}
\usepackage{multirow}
\usepackage{tabularx}
\usepackage{makecell}
\usepackage{enumitem}
\usepackage{siunitx} 
\newcolumntype{Y}{>{\raggedright\arraybackslash}X}
\newcolumntype{L}{>{\raggedright\arraybackslash}X}

\definecolor{codegreen}{rgb}{0,0.6,0}
\definecolor{codegray}{rgb}{0.5,0.5,0.5}
\definecolor{codepurple}{rgb}{0.58,0,0.82}
\definecolor{backcolour}{rgb}{0.95,0.95,0.92}

\lstdefinestyle{mystyle}{
    backgroundcolor=\color{backcolour},   
    commentstyle=\color{codepurple},
    keywordstyle=\color{magenta},
    numberstyle=\tiny\color{codegray},
    stringstyle=\color{codegreen},
    basicstyle=\ttfamily\footnotesize,
    breakatwhitespace=false,         
    breaklines=true,                 
    captionpos=b,                    
    keepspaces=true,                 
    numbers=left,                    
    numbersep=5pt,                  
    showspaces=false,                
    showstringspaces=false,
    showtabs=false,                  
    tabsize=2
}
\def\BibTeX{{\rm B\kern-.05em{\sc i\kern-.025em b}\kern-.08em
    T\kern-.1667em\lower.7ex\hbox{E}\kern-.125emX}}
\begin{document}

\title{SnapStream: Efficient Long Sequence Decoding on Dataflow Accelerators}


\author{%
\IEEEauthorblockN{%
Jonathan Li\IEEEauthorrefmark{1}\thanks{Correspondence to: jonathan.li@sambanovasystems.com},
Nasim Farahini\IEEEauthorrefmark{1},
Evgenii Iuliugin\IEEEauthorrefmark{1},
Magnus Vesterlund\IEEEauthorrefmark{1},
Christian Häggström\IEEEauthorrefmark{1},\\
Guangtao Wang\IEEEauthorrefmark{4},
Shubhangi Upasani\IEEEauthorrefmark{1},
Ayush Sachdeva\IEEEauthorrefmark{2},
Rui Li\IEEEauthorrefmark{1},
Faline Fu\IEEEauthorrefmark{1},\\
Chen Wu\IEEEauthorrefmark{1},
Ayesha Siddiqua\IEEEauthorrefmark{1},
John Long\IEEEauthorrefmark{1},
Tuowen Zhao\IEEEauthorrefmark{1},
Matheen Mussadiq\IEEEauthorrefmark{1},\\
Håkan Zeffer\IEEEauthorrefmark{1},
Yun Du\IEEEauthorrefmark{1},
Mingran Wang\IEEEauthorrefmark{1},
Qinghua Li\IEEEauthorrefmark{1},
Bo Li\IEEEauthorrefmark{1},\\
Urmish Thakker\IEEEauthorrefmark{3},
Raghu Prabhakar\IEEEauthorrefmark{1}}
\IEEEauthorblockA{\IEEEauthorrefmark{1}SambaNova Systems, Inc.}
\IEEEauthorblockA{\IEEEauthorrefmark{2}Cartesia AI}
\IEEEauthorblockA{\IEEEauthorrefmark{3}Microsoft AI}
\IEEEauthorblockA{\IEEEauthorrefmark{4}Meta Platforms, Inc.}
}

\maketitle

\begin{abstract}
The proliferation of 100B+ parameter Large Language Models (LLMs) with 100k+ context length support have resulted in increasing demands for on-chip memory to support large KV caches. Techniques such as StreamingLLM \cite{xiao2024efficientstreaminglanguagemodels} and SnapKV \cite{li2024snapkvllmknowslooking} demonstrate how to control KV cache size while maintaining model accuracy. Yet, these techniques are not commonly used within industrial deployments using frameworks like vLLM or SGLang. The reason is twofold: on one hand, the static graphs and continuous batching methodology employed by these frameworks \cite{kwon2023efficientmemorymanagementlarge,zheng2024sglangefficientexecutionstructured} make it difficult to admit modifications to the standard multi-head attention algorithm \cite{vllmstreamingllmsupport2023}, while on the other hand, the accuracy implications of such techniques on modern instruction-following and reasoning models are not well understood, obfuscating the need for implementing them. In this paper, we explore these accuracy implications on Llama-3.1-8B-Instruct and DeepSeek-R1, and develop  SnapStream, a KV cache compression method that can be deployed at scale. We demonstrate the efficacy of SnapStream in a 16-way tensor-parallel deployment of DeepSeek-671B on SambaNova SN40L accelerators running at 128k context length and up to 1832 tokens per second in a real production setting. SnapStream enables 4 times improved on-chip memory usage and introduces minimal accuracy degradation on LongBench-v2, AIME24 and LiveCodeBench. To the best of our knowledge, this is the first implementation of sparse KV attention techniques deployed in a production inference system with static graphs and continuous batching.
\end{abstract}

\begin{IEEEkeywords}
LLM serving, KV cache compression, dataflow architecture
\end{IEEEkeywords}

\section{Introduction}
Modern Large Language Models (LLMs) consistently use more than 100B parameters~\cite{deepseekr1,yang2025qwen3technicalreport,openai2025gptoss120bgptoss20bmodel,kimiteam2025kimik2openagentic}, requiring hundreds of gigabytes of on-chip memory to serve a single instance. At the same time, the rise of function-calling and test-time reasoning has significantly increased LLM input and output sequence lengths to 100k+ tokens, leading to large KV caches that exacerbate memory pressure.

\begin{figure}
    \centering
    \includegraphics[width=0.9\linewidth]{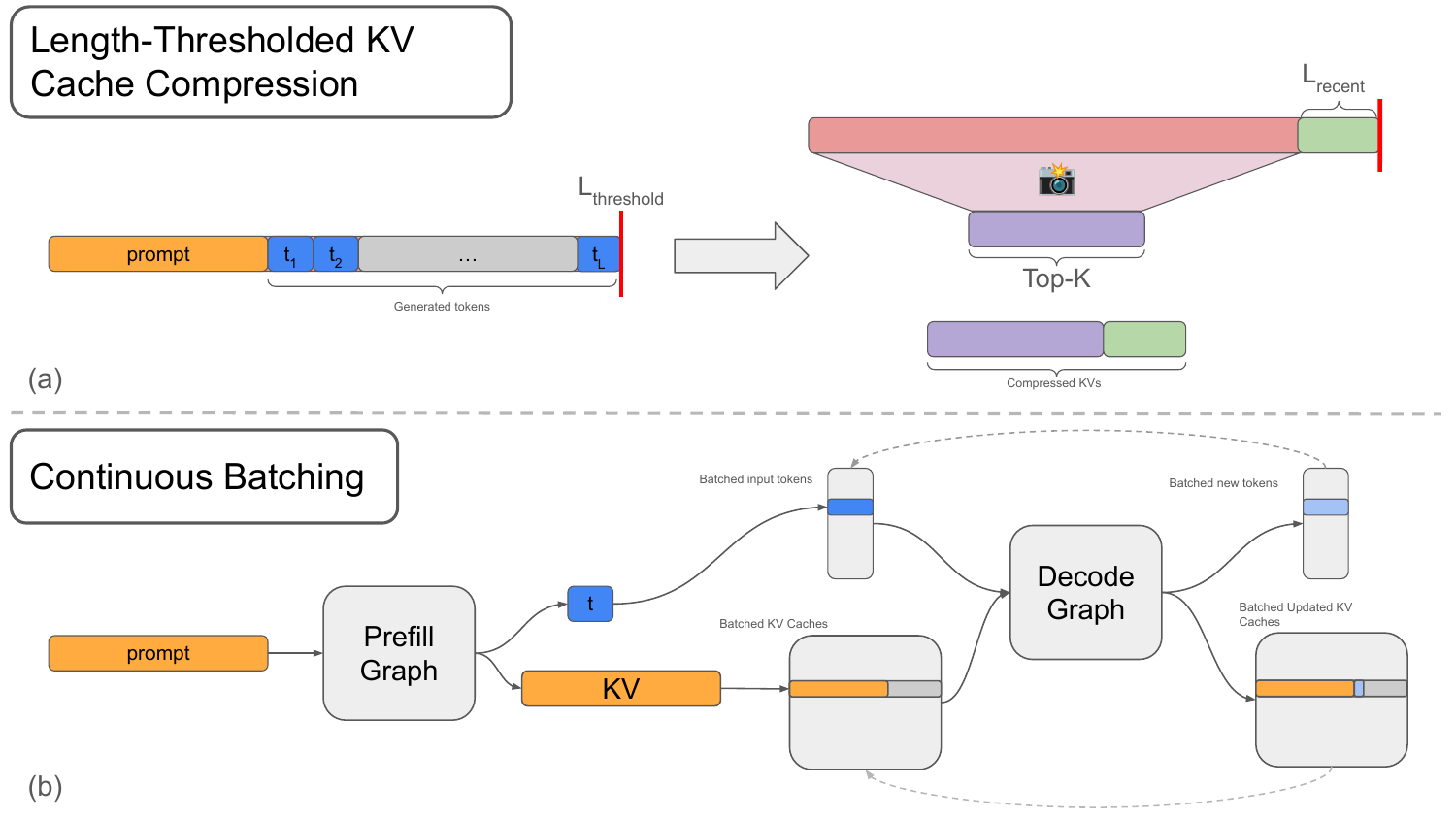}
    \caption{(a) Common KV cache compression methods like SnapKV \cite{li2024snapkvllmknowslooking} perform compression when the input sequence reaches length $L_\text{threshold}$. (b) Continuous Batching deployments consist of two graphs: a prefill graph that produces a single new token and the KV cache, and a decode graph that generates the next token and an updated KV cache. It's unclear where KV cache compression can be performed in this process, as the threshold $L_\text{threshold}$ can be reached during either prefill or decode and at different times for different batch elements.}
    \label{fig:kvcompress_and_cb}
\end{figure}

In response, practitioners have developed a variety of architectural modifications and compression methods. Multi-query attention \cite{ainslie2023gqatraininggeneralizedmultiquery}, multi-head latent attention \cite{deepseekai2024deepseekv2strongeconomicalefficient} and sliding window attention \cite{gemmateam2024gemma2improvingopen} each modify the attention operation to reduce memory requirements of the attention head, hidden and sequence dimensions respectively. However, these methods generally involve training a model from scratch with the proposed attention alternatives, reducing their applicability to existing models. KV cache compression methods, such as H$_2$O~\cite{zhang2023h2oheavyhitteroracleefficient} and SnapKV \cite{li2024snapkvllmknowslooking}, take advantage of high sparsity levels in the attention matrix~\cite{zhang2025spargeattentionaccuratetrainingfreesparse} to evict tokens with low attention scores from the KV cache. Such methods are largely training-free and can reduce KV cache memory by up to 92\% with negligible decrease in benchmark scores. However, when applying these techniques to modern LLM production deployments, the following practical considerations emerge:
\begin{itemize}
    \item \textbf{Long sequence decoding}: KV cache compression methods are generally evaluated on benchmarks with long inputs \cite{bai2024longbenchbilingualmultitaskbenchmark,niah} by compressing the inputs once at the beginning of decoding. The effects of KV cache compression are not well understood for reasoning models \cite{wei2023chainofthoughtpromptingelicitsreasoning} that may generate thousands of tokens for a single response, as shown in Table~\ref{tab:deepseek_tokens}.
    \item \textbf{Continuous batching}: Cloud LLM deployments typically use continuous batching \cite{zhong2024distservedisaggregatingprefilldecoding} to decouple the compute-bound prefill stage of LLM serving from the memory-bound decoding stage (see Fig.~\ref{fig:kvcompress_and_cb}(b) for a simplified workflow). It is not clear where KV cache compression fits into this workflow; existing implementations conditionally compress KV caches based on reaching some threshold sequence length, which can occur during either prefill or decoding, not to mention at different times for different batch elements.
    \item \textbf{Static tensor shapes}: Production LLM systems~\cite{kwon2023efficientmemorymanagementlarge,tensorrt,Prabhakar_2024} use compute graphs with fixed tensor shapes to better allocate on-chip memory resources. Existing KV cache compression methods are implemented with dynamic tensor shapes and tend to use slicing, indexing, and concatenation operations that often allocate small, short-lived buffers and promote fragmentation. 
\end{itemize}

In this paper, we introduce SnapStream, a model deployment strategy compatible with continuous batching (CB). SnapStream involves applying SnapKV~\cite{li2024snapkvllmknowslooking} cache compression during prefill and StreamingLLM~\cite{xiao2024efficientstreaminglanguagemodels} during decoding to generate long sequences with a smaller fixed KV cache size. To encourage adoption of this technique by frameworks that optimize tensor allocation with static graphs, we describe how to modify a static KV cache deployment with implementations of these techniques using fixed tensor sizes, and show an example mapping of these modifications to a 16-way tensor parallel (TP16) deployment on SN40L RDUs~\cite{Prabhakar_2024}. We derisk the accuracy of applying SnapKV and StreamingLLM together by benchmarking Llama 3.1 8B Instruct on LongBench, RULER, and $\infty$Bench and DeepSeek-R1 671B on LongBench-v2, AIME24, and LiveCodeBench. For DeepSeek-R1, our application of SnapStream in a TP16 deployment results in a $4\times$ reduction in KV cache memory and a $4.3\times$ improvement in decoding throughput, while incurring at most 5\% higher latency during prefill. SnapStream enables prefill and decoding for long sequences with significantly lower memory requirements than a standard disaggregated continuous batching (DCB) deployment, saving more memory with longer contexts and larger models. In summary, our contributions are as follows:
\begin{itemize}
    \item We introduce SnapStream, an efficient KV cache compression method designed for LLM serving at scale with continuous batching deployments. We describe the structure of the compressed KV cache, and the modifications that must be made to prefill and decoding phases to implement SnapStream.
    \item We describe kernel fusion and tensor sharding strategies for a static graph deployment of DeepSeek-R1 on a SN40L-16 node. We show how different mapping strategies for prefill and decoding optimize for different service level objectives (SLOs), and how SnapStream can be accommodated by adding fused kernels to the prefill execution schedule.
    \item We show that SnapStream results in minimal accuracy degradation by benchmarking Llama 3.1 8B Instruct and DeepSeek-R1-0528 on both long sequence benchmarks and reasoning benchmarks.
    \item We demonstrate the efficacy of SnapStream by measuring  $4\times$ higher maximum batch size and $4.3\times$ higher throughput during decoding when applied to a TP16 deployment of DeepSeek-R1.
\end{itemize}

\section{Preliminaries}

\subsection{Continuous Batching}
LLM serving is defined by two stages of computation: the prefill phase and the decode phase, each with their own service level objectives (SLOs). These SLOs are time to first token (TTFT) and time per output token (TPOT) for prefill and decoding respectively. In prefill, the LLM processes a user prompt of length $L$ to generate a new token $x_{L+1}$ and KV cache tensors of length $L$. In this phase, the $O(L^2)$ attention mechanism dominates latency, making the prefill phase compute bound. In the decoding phase, the model processes a KV cache of length $L$ and a single input token $x_{L+1}$ to generate the next token $x_{L+2}$ and the KV cache of length $L+1$. By re-using the key and value tensors in the KV cache, multi-head attention becomes $O(L)$ during decoding. With significantly lower computational intensity compared to the prefill phase, the decode phase is typically memory bound. In practice, compute-intensive long sequence prefill requests are often served with low batch size to minimize TTFT, while memory-intensive decoding requests are served with higher batch size to maximize hardware utilization.

Continuous batching (CB) is a popular LLM serving technique \cite{Yu2022OrcaAD,agrawal2023sarathiefficientllminference,kwon2023efficientmemorymanagementlarge} in which batches of prefill and decoding requests may be interleaved on the same system, and the output of completed prefill requests can be added to decode batches to improve utilization. However, there is an inherent contention to the different SLOs for prefill and decoding, as indicated by the difference between their batching preferences. Disaggregated Continuous Batching (DCB) \cite{zhong2024distservedisaggregatingprefilldecoding} proposes to resolve this contention by assigning separate hardware resources to prefill and decoding phases and transferring KV caches from prefill nodes to decoding nodes between phases. The authors show that DCB can sustain up to a $7.4\times$ request rate compared to the leading CB baselines on a 32 GPU, 4 node setup.

\subsection{Reconfigurable Dataflow Units (RDUs)}

\begin{figure}
        \centering
         \includegraphics[width=\linewidth]{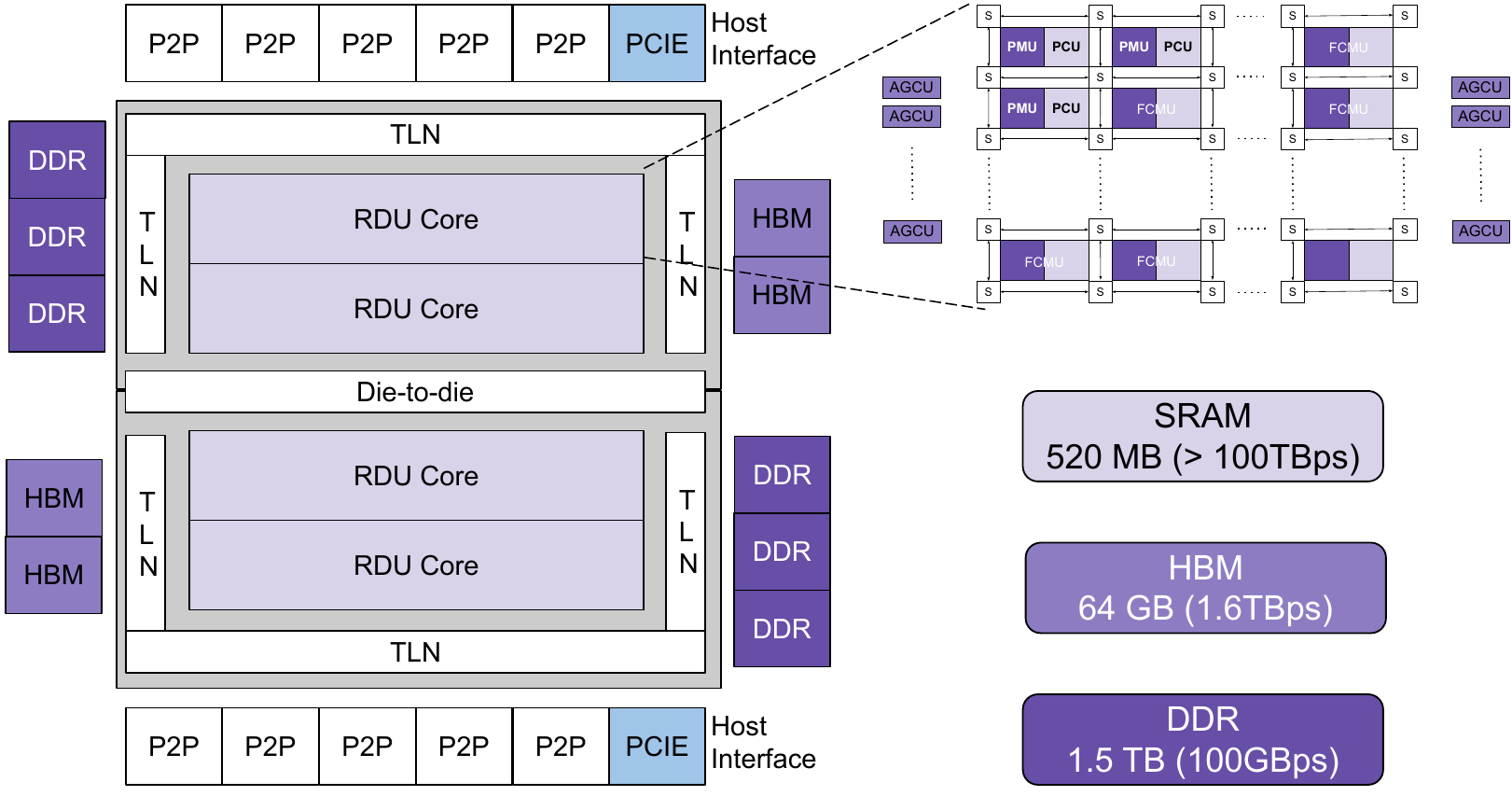}
         \caption{SN40L Architecture. Packaged as a two-die socket in 5FF TSMC process. Each die features 2 dense compute Tiles, 2 HBM modules, and 3 DDR channels. Tiles are interconnected via the Top Level Network (TLN) and can communicate with other RDUs using the P2P interfaces. Each Tile is comprised of PCUs and PMUs connected in a mesh network, RDN, enabling seamless data exchange.}
         \label{fig:sn40l}
\end{figure}

Built on a dataflow architecture, the SN40L AI accelerator optimizes AI training and inference tasks. Prior works~\cite{Prabhakar_2024,sn40l-hotchips} describe the architecture in more detail. This section provides a brief overview of key architectural components. Each socket features four RDU cores, I/O blocks for connecting to DDR, high bandwidth memory (HBM), Peer-to-Peer (P2P) links, and a host system (see Fig.~\ref{fig:sn40l}). A single socket can perform 638 TFLOPs of BF16 operations and has 520 MB of on-chip SRAM. Each socket also interfaces with 64 GB of HBM memory providing 1.6 TB/s of peak bandwidth, and 1.5 TB of DDR memory providing over 100 GB/s of peak bandwidth. A Top-Level Network (TLN) connects tile components to the IO blocks. Each socket consists of 1040 Pattern Compute Units (PCUs), which provide systolic and streaming compute capabilities, and 1040 Pattern Memory Units (PMUs) which provide program-managed on-chip memory for storing tensors. Collectively, a PCU and PMU block is known as a Fused Compute Memory Unit (FCMU). Additionally, the Address Generation and Coalescing Units (AGCUs) serve as the interface to external memory and other sockets.

\subsection{LLM Serving with Static Graphs}
\label{subsec:static_llm_serving}
The SN40L compiler operates primarily on static graphs, where all operations are executed on tensors with fixed shapes. For an LLM, all operations are executed with a fixed sequence length $L_{\text{max}}$ such that, given an input token sequence $x_1,...,x_L$ with $L \leq L_{\text{max}}$, we operate on the padded sequence $x_1,...,x_L,x_{\text{pad}:L+1},...,x_{\text{pad}:L_{\text{max}}}$. Executing the prefill graph generates a single new token $x_{L+1}$ as well as key and value sequences $k^{l,h}_1,...,k^{l,h}_L,k^{l,h}_{\text{pad}:L+1},...,k^{l,h}_{\text{pad}:L_{\text{max}}}$ and $v^{l,h}_1,...,v^{l,h}_L,v^{l,h}_{\text{pad}:L+1},...,v^{l,h}_{\text{pad}:L_{\text{max}}}$ for the $l$\textsuperscript{th} hidden layer and $h$\textsuperscript{th} attention head. \footnote{For brevity, in future references to these variables we will omit the layer and head indices $l,h$. Similarly, we will only describe operations on the key values $k$, as our treatment of the value tensors $v$ is mostly identical.} Executing the decode graph immediately after prefill generates new key and value tensors $k_{L+1}$ and $v_{L+1}$ that are used to update the KV cache \textit{in-place} (i.e. the new output key sequence looks like $k_1,...,k_L,k_{L+1},k_{\text{pad}:L+2}...,k_{\text{pad}:L_{\text{max}}}$) and generates one more new token $x_{L+2}$. The new token and the updated KV cache are used as input for another call to the decode graph. This process is repeated until the \texttt{EOS} token or $x_{L_{\text{max}}}$ is generated. See Fig.~\ref{fig:snapstream} (a) and (c) for a visualization of this process.

\subsection{SnapKV}
\label{subsec:snapkv}

SnapKV is a KV cache compression method that mitigates the lossy compression seen in prior compression schemes by using a pooling-based clustering mechanism. The pooling layer helps to retain critical token clusters with higher attention weights, thus retaining the completeness of information. This clustering is particularly beneficial in retrieval tasks like needle-in-a-haystack, as we later show in the Retrieval subset of RULER in Table~\ref{tab:llama_acc}. Formally, given a prompt with length $L = L_{\text{prefix}} + L_{\text{obs}}$ divided into a prefix and an observation window, for the $l$\textsuperscript{th} hidden layer we obtain the SnapKV index set $I_l$ by computing:
\begin{align}
    C_l &= \sum_{i=0}^{L_{\text{obs}}} W_l[:, :, i, :]\label{eq:snapkv_summed_attn} \\
    I_l &= \text{Top}_K(\text{avgpool}(C_l)) \label{eq:snapkv_topk}
\end{align}
where $W_l\in\mathbb{R}^{B\times H\times L_{\text{obs}}\times L_{\text{prefix}}}$ is the softmax-normalized attention distribution of tokens in the observation window attending to prefix tokens for layer $l$ with batch size $B$ and number of attention heads $H$.\footnote{As with KV sequences, in future references we will omit the layer index $l$, as operations are identical across layers.}

\subsection{StreamingLLM}
StreamingLLM~\cite{xiao2024efficientstreaminglanguagemodels} is a technique that enables LLMs to process sequences beyond their fixed context windows, effectively supporting unbounded streaming inputs and outputs. Conventional LLMs are limited by a fixed attention window, which constrains their ability to maintain coherence over long sequences. StreamingLLM addresses this by retaining a small set of attention sink tokens (the initial prompt tokens that consistently receive high attention across decoding steps) together with the most recent tokens. These tokens act as stable anchors that preserve long-term context, allowing the model to maintain continuity without storing the entire KV cache.

The core mechanism of StreamingLLM is a compact KV cache strategy that combines these sink and recent tokens to form a reduced attention context. During decoding, the model attends only to this subset, substantially lowering memory usage while preserving model accuracy.

\section{Methodology}
\begin{figure}
    \centering
    \includegraphics[width=0.9\linewidth]{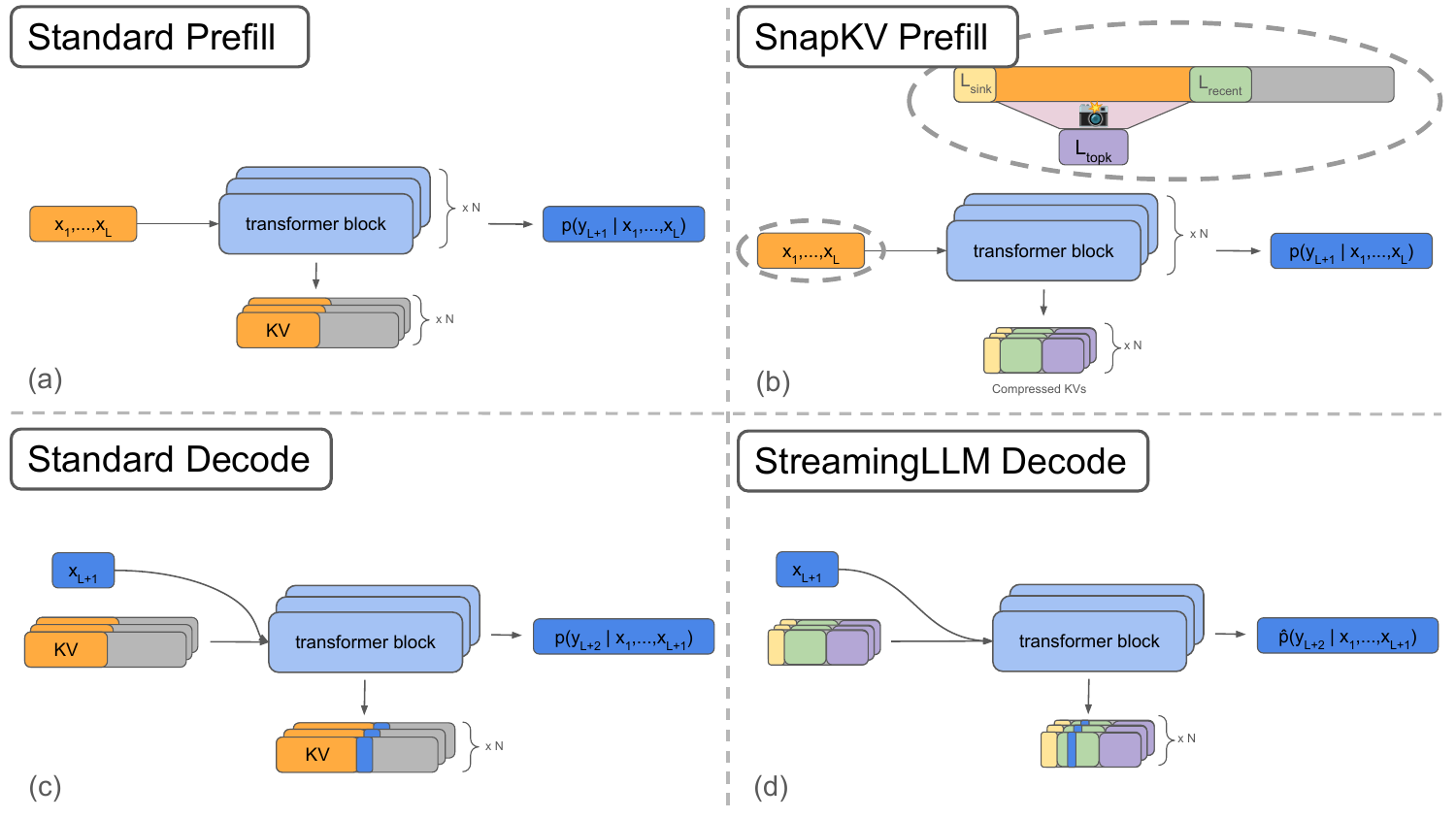}
    \caption{SnapStream  applies SnapKV during prefill (b) to produce a compressed KV cache and StreamingLLM during decoding (d) to update the recent tokens of the compressed cache in-place. In contrast, standard static graph prefill (a) produces a padded KV cache that is appended to during decoding (c).}
    \label{fig:snapstream}
\end{figure}

SnapStream is built on top of a static graph CB deployment by applying SnapKV to prefill KV caches and decoding with StreamingLLM. Following~\cite{xiao2024efficientstreaminglanguagemodels}, given an input sequence of length $L$, during prefill we extract the first $L_{\text{sink}}$ tokens and the last $L_{\text{recent}}$ tokens from the full KV cache. We apply SnapKV compression to the remaining $L - (L_\text{sink} + L_{\text{recent}})$ tokens, extracting the top $K$ tokens. We concatenate these three tensors along the seqeuence dimension to obtain a prefill KV cache of length $L_{\text{snapstream}} = L_{\text{sink}} + L_{\text{recent}} + K$. During decoding, we decode with a fixed KV cache size of $L_{\text{snapstream}}$, treating the middle $L_{\text{recent}}$ tokens as the rotating KV cache from StreamingLLM. See Fig.~\ref{fig:snapstream}(b) and (d) for a visualization of SnapStream prefill and decoding, respectively. In the following sections, we describe the layout of the SnapStream KV cache and the operation of SnapStream prefill and decoding in a production static-graph execution framework.

\subsection{KV Cache Structure}
\label{subsec:kvcachestructure}
In SnapStream, the KV cache has three distinct components, indexed by their sequence position relative to the prefill length $L$:
\begin{itemize}
    \item Sink tensors $k_1,...,k_{L_{\text{sink}}}$
    \item Recent tensors $k_{L - L_{\text{recent}}},...,k_L$
    \item Top-$K$ tensors $k_{\text{topk}:1},...,k_{\text{topk}:K}$
\end{itemize}

See Fig.~\ref{fig:snapstream}(b) for a depiction of the three components in yellow, green, and purple, respectively. Each component is initialized during prefill, with top-$K$ tensors appended to the sink and recent tensors after any padding. Let $L_{sr} = L_{\text{sink}} + L_{\text{recent}}$; then, the exact structure of the KV cache has two cases:

\subsubsection{Case 1: $L < L_{sr}$}
In this case, the key tensor cache is constructed during prefill as $k_1,...,k_L,k_{\text{pad}:L+1},..,k_{\text{pad}:L_{sr}},k_{\text{topk}:1},..,k_{\text{topk}:K}$. This structure is no different than that described in Section~\ref{subsec:static_llm_serving}, as the Top-$K$ tokens $x_{\text{topk}:\cdot}$ are themselves padding tensors in this case. Accordingly, during decoding, the KV cache is updated in-place in the same manner as previously described in Section~\ref{subsec:static_llm_serving}. 

\subsubsection{Case 2: $L \geq L_{sr}$}
In this case, the key cache is constructed as $k_{1},...,k_{L_{\text{sink}}},k_{L - L_{\text{recent}}},...,k_L,k_{\text{topk}:1},...,k_{\text{topk}:K}$. During prefill, we remove the middle $L - L_{sr}$ tokens from the KV cache, and extract the top $K$ key and value tensors from these tokens to popuate $k_{\text{topk}:1},...,k_{\text{topk}:K}$. During decoding, we maintain the same KV cache size by evicting the least recently used key tensor, $k_{L - L_{\text{recent}}}$, and replacing it with the new key tensor $k_{L+1}$. Note that in this case, the sink tensors and top-$K$ tensors are never modified during decoding.

\subsection{Prefill}
\label{subsec:prefill}
In order to reduce memory usage of the decoding stage, SnapStream applies SnapKV compression during prefill to compress an output KV cache along the sequence dimension. Adherent to the KV cache structure described in Section~\ref{subsec:kvcachestructure}, we only apply SnapKV to compress the evicted tokens $k_{L_{\text{sink}}+1},..., k_{L - L_{\text{recent}}}$. The sink token KV cache, recent token KV cache, and compressed SnapKV cache are then concatenated and transferred to the decoding node. 

In preparing a KV cache for StreamingLLM decoding, we make two modifications to vanilla StreamingLLM during prefill:
\begin{itemize}
    \item We apply StreamingLLM \textit{after} any positional encodings are applied to query, key, and value projections. This means that we do \textit{not} apply positional encodings to cached KVs during decoding and use the original relative positional encodings. Although this may affect the ability of the model to generate beyond its trained attention window size as claimed in Section~3.2 of the StreamingLLM manuscript~\cite{xiao2024efficientstreaminglanguagemodels}, we believe that this change is well-motivated, as most modern open models are now trained to handle context lengths up to 128k~\cite{grattafiori2024llama3herdmodels,deepseekai2025deepseekv3technicalreport,yang2025qwen3technicalreport}, which is more than sufficient for most use cases.
    \item We implement the rolling KV cache as a ring buffer, as opposed to a sliding window constructed with repeated  concatenations of tensor views.
\end{itemize}

\subsubsection{StreamingLLM Ring Buffer}
\begin{figure}
    \centering
    \includegraphics[width=0.9\linewidth]{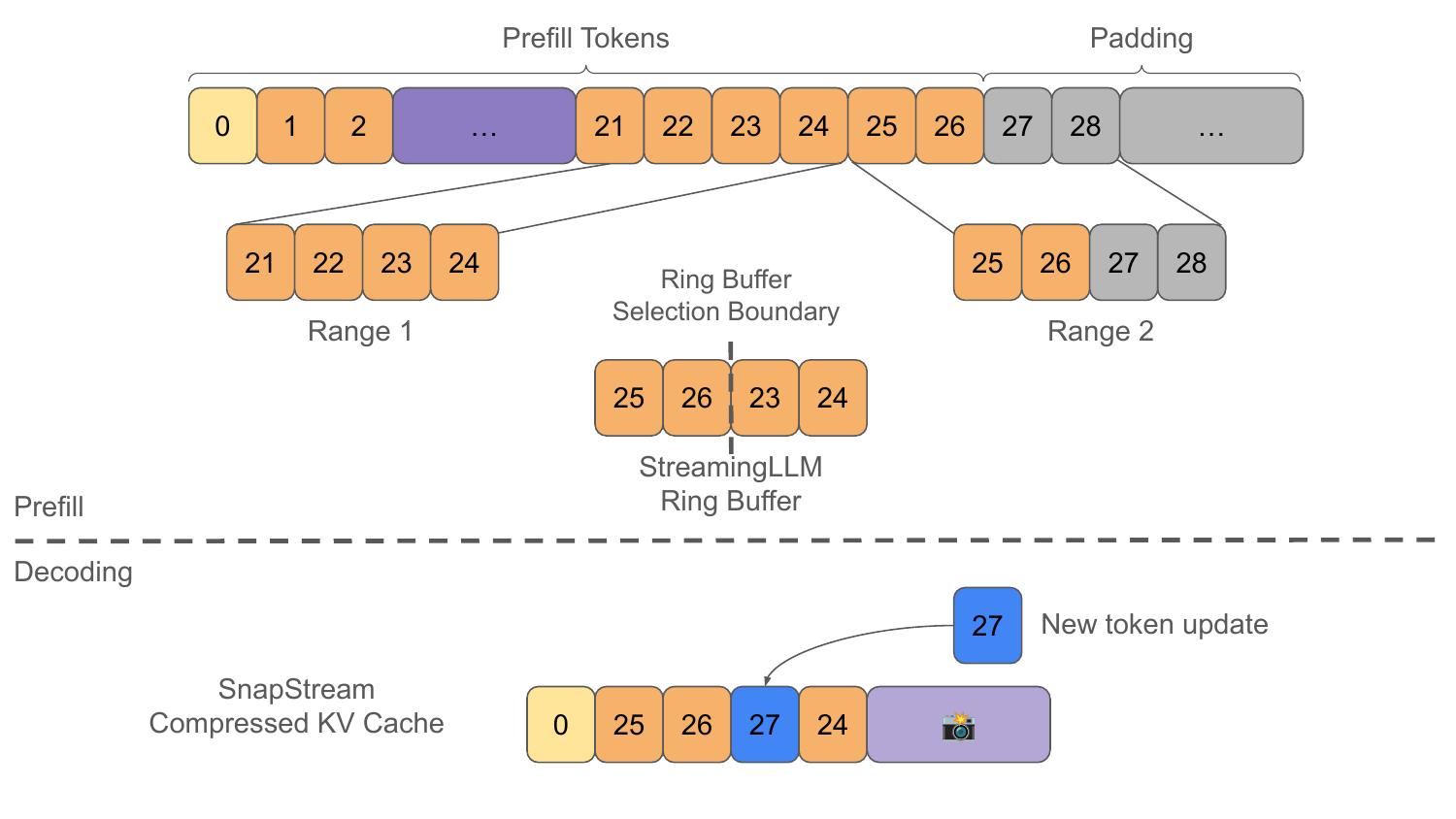}
    \caption{An example of how the SnapStream ring buffer is constructed during prefill, and how it is updated during decoding. See Listing~\ref{lst:prefillcompression} in the Appendix for prefill pseudocode. Given an input sequence with $L=26$, $L_{\text{sink}} = 1$, $L_{\text{recent}} = 4$, we gather KVs from indices 21-24 as Range 1 and 25-28 as Range 2. The ring buffer is constructed with indices 0-1 from Range 2 and indices 2-3 from Range 1. During decoding, we replace the KV for token index 23 with the KV for the newly generated index 27.}
    \label{fig:streamingllmranges}
\end{figure}

In this section, we motivate and describe the implementation of the StreamingLLM rolling KV cache window, which is implemented as a ring buffer. For a na\"ive implementation of StreamingLLM using a static KV cache size $L_{\text{stream}}$, generating beyond $L_{\text{stream}}$ tokens one at a time involves inefficient slicing of the sink tokens, recent token window, and top-$k$ tokens from the KV cache and concatenating them all together with the new keys and values (see Lines 24-32 of Listing~\ref{lst:naivedecodeupdate} in the Appendix). When the rolling window is implemented as a ring buffer, we only need to scatter new keys and values into the appropriate index during decoding (see Fig.~\ref{fig:snapstream}(d)). This requires us to prepare the ring buffer structure during the prefill phase, which involves two \verb!gather! calls for the sequence indices before and after the prefill length index. See Listing~\ref{lst:prefillringbuffer}  in the Appendix for a PyTorch-style implementation of the ring buffer construction, and Fig.~\ref{fig:streamingllmranges} for a visualization of where the two gather calls are placed along the sequence dimension.

\subsubsection{Recomputing Attention for SnapKV}
\label{subsubsec:snapkv_recompute}
In implementing SnapKV, we generally follow the original procedure outlined in Section~\ref{subsec:snapkv}, except that eviction candidates in SnapStream exclude sink tokens. We define $Q_{\text{obs}} = [q_{L-L_{\text{obs}}},...,q_L]$ as the query projections of the most recent $L_{\text{obs}}$ tokens and $K_{\text{evict}} = [k_{L_{\text{sink}}+1},...,k_{L-L_{\text{recent}}}]$ as the key projections of the set of eviction candidates. Then, the top-$K$ indices $I$ are computed exactly as in Equations~\ref{eq:snapkv_summed_attn} and~\ref{eq:snapkv_topk}, except that $W\in\mathbb{R}^{B\times H\times L_{\text{obs}}\times L_{\text{evict}}}$ is defined as
\begin{align}
    W &= \text{softmax}\left(\frac{Q_{\text{obs}}^\intercal K_{\text{evict}}}{\sqrt{d}}\right) \label{eq:snapattn}
\end{align}

We note that FlashAttention~\cite{dao2022flashattentionfastmemoryefficientexact} and other optimized attention implementations generally do not give their users access to the $O(L^2)$ attention matrix $QK^\intercal$, as much of their value proposition comes from avoiding the expensive procedure of writing the attention matrix from SRAM to HBM. This has led to difficulties implementing sparse attention and efficient KV cache methods in libraries like vLLM~\cite{vllmstreamingllmsupport2023}. To get around this issue, our implementation recomputes Equation~\ref{eq:snapattn} outside of the fused attention kernel during prefill (see Section~\ref{subsec:prefill}); we find that this amount of recompute constitutes less than 3\% of the total prefill latency, but future works may optimize this step to avoid this additional compute.

Once the sink tokens and the recent token ring buffer are constructed, the Top-$K$ KVs are gathered from the removed tokens using SnapKV and are appended at the end. Since the input sequence length $L$ may be less than $L_{sr}$, the Top-$K$ tokens may consist of some number of padding tokens $0 \leq p\leq K$. We pass the number of padding tokens for each sequence as additional metadata from the prefill stage to decoding to be used for the construction of an attention mask that masks out any padding tokens in the Top-$K$ section of the SnapStream KV cache. 

\subsection{Decoding}
By constructing the KV cache as a ring buffer, the decoding stage of a SnapStream deployment remains almost exactly the same as the standard decoding stage using static graphs described in Section~\ref{subsec:static_llm_serving}, except that the in-place KV updates are performed at the modified position $L_{rb}$ defined in Equation \ref{eq:lti}. This means that we do not have to introduce any additional buffers from tensor indexing and concatenation that may be introduced by a na\"ive implementation of StreamingLLM (see Listing~\ref{lst:naivedecodeupdate} in the Appendix). In particular, any fused kernels developed for decoding, such as those described in Section~\ref{sec:prefill-mapping}, may be used without modification.
\begin{align}
    L_{rb} = (L - L_\text{sink} + L_\text{recent}) \mod L_{\text{recent}} + L_\text{sink}
    \label{eq:lti}
\end{align}

\section{Implementation}
\label{subsec:mapping}

Our goal in formulating SnapStream is to bridge the implementation gap between KV cache compression and LLM deployments, particularly for frameworks that use static graphs. We demonstrate the effectiveness of our approach by applying SnapStream to a real deployment of a state-of-the-art (SoTA) reasoning LLM, DeepSeek-R1-0528~\cite{deepseekr1}, on the SN40L accelerator. In this section, we describe the kernel fusion and tensor sharding decisions, known as a ``mapping'', for a single decoder layer of DeepSeek-R1-0528 on a single SN40L-16 node. Separate mappings are created for prefill and decoding that optimize for each phase's respective SLOs. We also show how SnapStream integrates seamlessly into these mappings.

\subsection{Prefill}
\label{sec:prefill-mapping}

\begin{figure}
    \centering
    \includegraphics[width=1\linewidth]{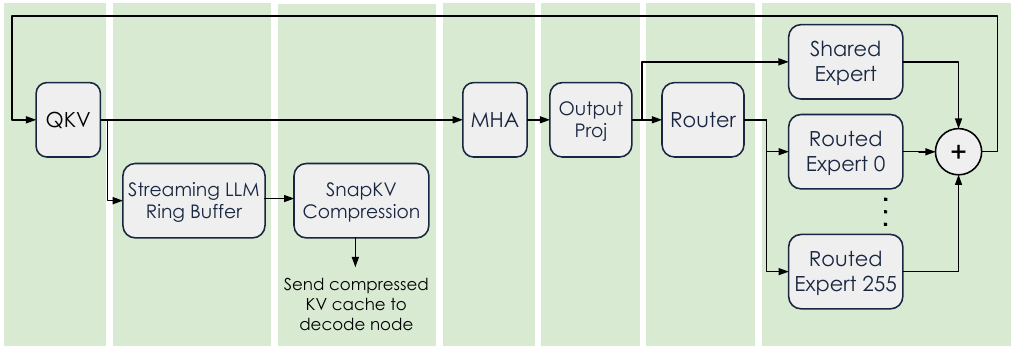}
    \caption{High-level block diagram of the modified MoE prefill graph incorporating SnapStream compression. The graph is decomposed into multiple fused kernels, indicated by green boxes.}
    \label{fig:deepseek-prefill}
\end{figure}

\begin{figure}
    \centering
    \includegraphics[width=1\linewidth]{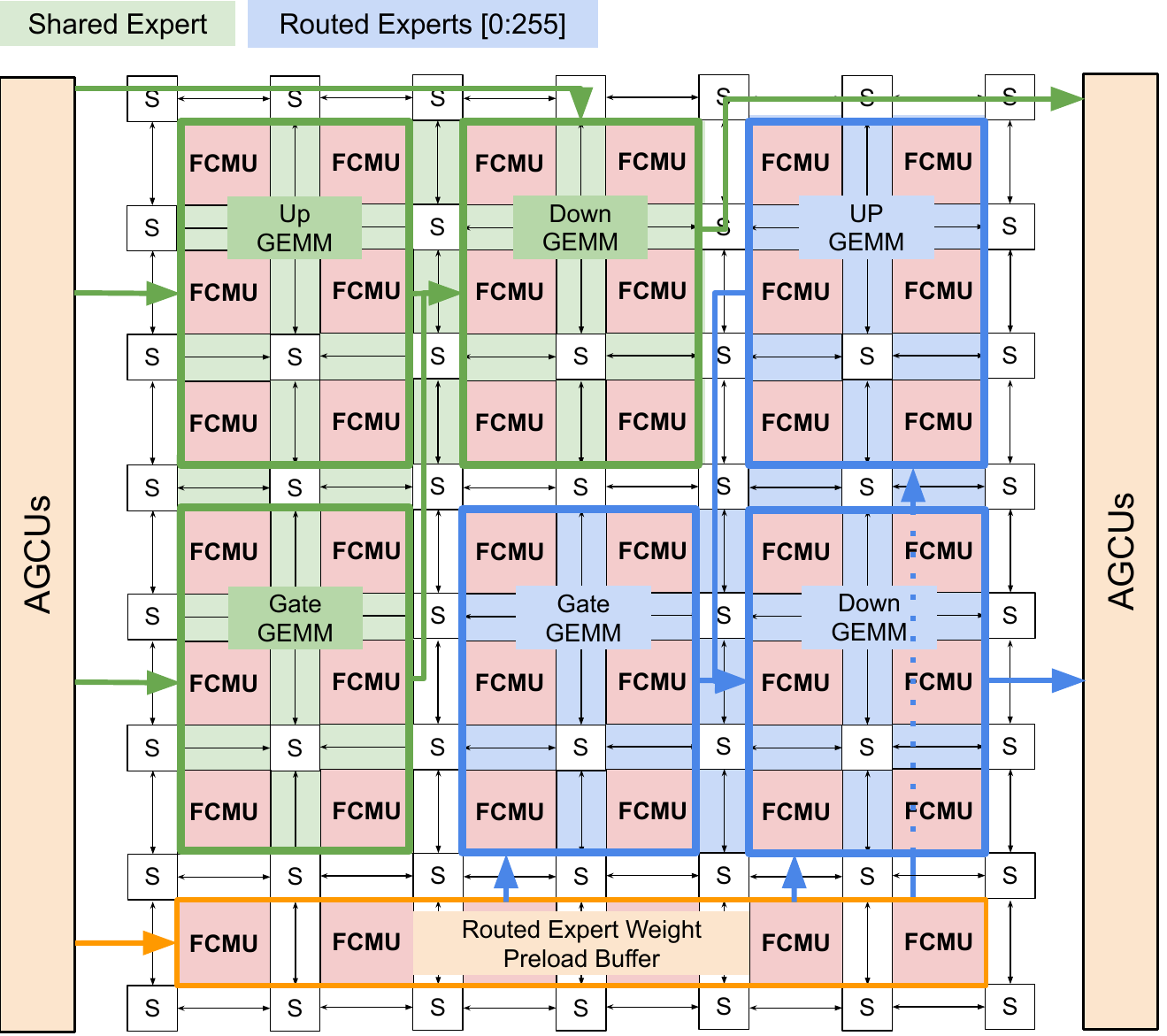}
    \caption{Spatially pipelined and fused implementation of MoE FFN. Data is chunked and streamed through operators in the fused kernel, allowing early initiation of P2P communication across sockets under TP16 partitioning and overlapping data transfer with computation.}
    \label{fig:ffn_on_rdu}
\end{figure}

The minimum prefill deployment unit consists of a single SN40L-16 node executing with 16-way tensor parallelism (TP16). Prefill runs at batch size 1 with continuous batching. DeepSeek-R1-0528 consists of two types of decoder layers: dense FFN and sparse Mixture-of-Experts (MoE) layers. Fig.~\ref{fig:deepseek-prefill} shows a high-level block diagram of the MoE prefill graph, including operators for SnapStream compression. The prefill graph is split into multiple fused kernels, with green boxes indicating kernel boundaries. The use of dataflow, large on-chip SRAM buffers, and a programmable interconnect enables unprecedented levels of operator fusion and pipelining. This results in increased operational intensity and reduced HBM traffic.

The QKV kernel fuses the layer normalization, skip-add, and the QKV projection GEMMs. Normally, the KV cache is generated by the this kernel. To accommodate SnapStream in prefill, we add two additional kernels to handle the construction of the StreamingLLM ring buffer and the SnapKV compression, respectively:


\begin{itemize}
    \item \textbf{StreamingLLM Ring Buffer}: this kernel fuses four gather operations for Range 1 and Range 2 from Fig.~\ref{fig:streamingllmranges} for each of the key and value caches, as well as indexing for the sink tensors. All operations are equally sharded across the number attention heads $H$ into 16 sockets.
    \item \textbf{SnapKV Compression}: this kernel fuses several operations described below. All operations are sharded across the number of attention heads $H$. The operations are:
    \begin{enumerate}[label=(\roman*)] 
        \item the recomputed $QK^\intercal$ GEMM described in Section~\ref{subsubsec:snapkv_recompute}
        \item the attention aggregation operations in Equation~\ref{eq:snapkv_summed_attn} and Equation~\ref{eq:snapkv_topk}
        \item two gather operations using the top-$K$ indices $I$ for each of the key and value tensors
        \item two concatenation operators joining the StreamingLLM ring buffer with the SnapKV-compressed key and value caches
    \end{enumerate}
\end{itemize}

Following the QKV, StreamingLLM Ring Buffer and SnapKV Compression kernels, the MHA kernel implements multi-head attention using the QKV projections of the previous sections. Similarly to FlashAttention, we fuse the entire $QK^\intercal$ multiplication, softmax and $PV$ multiplication into a single kernel to avoid materializing the full attention matrix $P$. We distribute all operators in the MHA kernel along the number of attention heads $H$ equally across 16 sockets, and tile $K$ and $V$ into blocks of 8192 within each socket.

The Output Proj kernel executes only the output projection GEMM. The Router kernel performs the router gate GEMMs, Top-$K$ expert sorting (a different $K$ than that used for SnapKV), and the selection of the highest-scoring experts for each token in the input sequence.

The entire MoE FFN - including the shared and routed expert operators as well as the activation function - is fused into a single kernel. This allows all dynamically selected experts to execute without synchronization overhead at kernel boundaries and to deliver substantial performance improvements. The fused implementation also enables preloading the weights of the next expert while processing the current one, further reducing the Time to First Token (TTFT) in prefill.

Fig.~\ref{fig:ffn_on_rdu} illustrates a spatially mapped implementation of the MoE FFN on SN40. Using spatial dataflow, we execute operators as a coarse-grained pipeline, where tensors are divided into tiles and streamed through sequential operators. Using TP16, each expert in a socket holds one-sixteenth of the weight matrix, requiring peer-to-peer (P2P) communication to gather activation fragments from all sockets before performing the down GEMM. The fused, pipelined, and tiled implementation enables P2P transfers to begin as soon as the first tile computation completes, effectively overlapping communication and computation, and further improving efficiency.

While we only describe the mapping for MoE layers here, the kernel boundaries for the dense layers are largely the same except that the Output Proj and FFN operator groups are fused into a single kernel.

\begin{figure}
    \centering
    \includegraphics[width=1\linewidth]{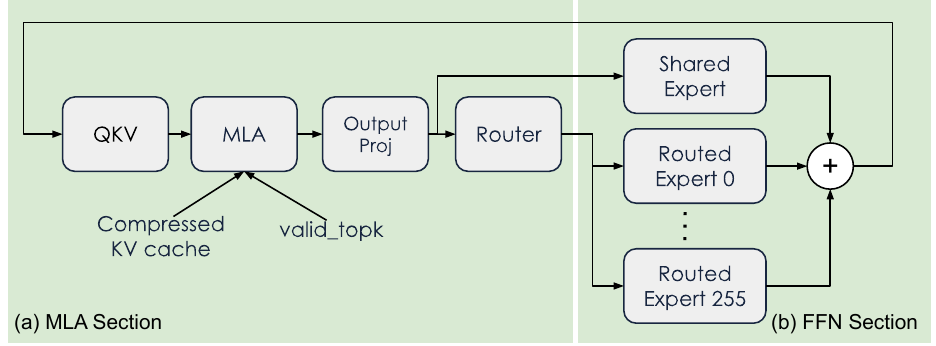}
    \caption{High-level block diagram of the decode computation graph operating on the compressed KV cache. The graph is split into two main sections, which are each compiled into a single fused kernel: (a) Multi-Head Latent Attention (MLA), including QKV projections, attention, output projection, and router GEMM; and (b) Feed-Forward Network (FFN), comprising shared and routed expert GEMMs.}
    \label{fig:deepseek-decode}
\end{figure}

\subsection{Decoding}
\label{sec:decode-mapping}

The decoding stage runs on a single SN40L-16 node. KV cache reuse significantly reduces the amount of computation required for decoding, effectively reducing the sequence dimension of most operations to 1. This allows us to compile even larger fused kernels for this stage, such that the entire computation graph is contained within two kernels: (a) a Multi-Head Latent Attention (MLA) kernel and (b) a Feed-Forward Network (FFN) kernel, as illustrated in Fig.~\ref{fig:deepseek-decode}.

The MLA kernel includes QKV projections, attention computation, output projection, and the MoE router GEMM. The FFN section contains the shared and routed expert GEMMs. Both sections are implemented using fusion and spatial dataflow, as shown in Fig.~\ref{fig:ffn_on_rdu}. 

Within the MLA section, the MLA operator runs using 16-way data parallelism (DP16) across the batch dimension, whereas QKV and output projection GEMMs use 16-way tensor parallelism (TP16). This configuration provides a balanced trade-off between memory capacity, communication overhead, and compute efficiency across the RDU sockets.

During decoding, the MLA operator does not benefit from tensor parallelism because splitting across attention heads would require every socket to access the full KV-cache for all batch samples. This results in inefficient memory capacity and bandwidth utilization, especially for large batch sizes. Instead, with DP16, the KV-cache is partitioned across the 16 sockets. This, combined with SnapKV compression, enables substantially larger batch sizes yielding higher throughput. Running QKV and output projection under TP16 distributes their weights across all sockets, further reducing the memory footprint and memory bandwidth requirements of the MLA section.

The FFN section is executed under TP16, allowing weight reuse across all batch samples. Each socket computes all samples, but only on 1/16th of the routed expert slice, and only stores 1/16th of the expert weights locally. This enables grouping of tokens by expert, where all tokens requiring the same expert are processed together. During decoding, we employ the batch–sample fusion technique ~\cite{nandkar2025speculativedecoding}, where samples assigned to the same expert are fused and assigned to adjacent PCU systolic stages to increase FLOP utilization. This approach ensures that the weights for each expert are loaded only once across all 16 sockets and that the hardware achieves high compute efficiency through reuse and fused execution.

Accommodating a SnapStream KV cache in the decoding graph is relatively simple, since all we need to change is the index used to do the KV cache update. This index is exactly the value $L_{rb}$ defined in Equation~\ref{eq:lti}, and its computation easily fits into the fused kernel of the MLA section.

\section{Experiments}
\label{sec:experiments}
\subsection{Accuracy}
We initially validate the accuracy of SnapStream by evaluating Llama 3.1 8B Instruct \cite{grattafiori2024llama3herdmodels} on three long sequence evaluation suites: LongBench~\cite{bai2024longbenchbilingualmultitaskbenchmark}, RULER~\cite{hsieh2024rulerwhatsrealcontext} and $\infty$Bench~\cite{zhang2024inftybenchextendinglongcontext}. The results in Table~\ref{tab:llama_acc} demonstrate that aggressive compression of the KV cache, from 128k to 8k, leads to minimal accuracy degradation. In most task categories, the SnapStream deployment matches or exceeds the baseline. We also compare against StreamingLLM~\cite{xiao2024efficientstreaminglanguagemodels} and SAGE-KV \cite{wang2025llmsknowdropselfattention} deployments. SAGE-KV is another attention-guided KV cache compression mechanism. Compared to SnapKV, SAGE-KV only uses the last token's attention distribution to guide KV cache eviction and eschews SnapKV's pooling operation. SnapStream outperforms or matches both techniques on most benchmarks; in particular, StreamingLLM results in significant accuracy degradations in RULER's Retrieval and Multi-Hop (MH) Tracing tasks, as well as all subtasks in $\infty$-Bench, whereas SAGE-KV fails the KV retrieval task in $\infty$-Bench. SnapStream achieves scores that are most similar to the full attention baseline in these tasks, while also scoring the highest in LongBench.

\begin{table}[t]
  \centering
  \caption{Benchmark results for Llama 3.1 8B Instruct on long sequence tasks.}
  \label{tab:llama_acc}
  \setlength{\tabcolsep}{2.5pt}      
  \renewcommand{\arraystretch}{1.06} 
  \scriptsize                        

  \begin{tabularx}{\columnwidth}{@{} L *{4}{c} @{}}
    \toprule
    \textbf{Benchmark} &
    \makecell{Llama 3.1\\8B Instr.} &
    \makecell{+ StreamingLLM\\8k} &
    \makecell{+ SAGE-KV\\8k} &
    \makecell{+ SnapStream\\8k}\\
    \midrule

    \multicolumn{5}{@{}l}{\textbf{LongBench}}\\
    Single-Doc QA (F1) & 48.37 & 47.69 & 48.16 & \textbf{48.59} \\
    Multi-Doc QA (F1)  & 40.37 & 41.05 & 39.97 & \textbf{41.18}\\
    Summ. (Rouge-L)    & \textbf{25.65} & \textbf{25.65} & 25.42 & \textbf{25.65} \\
    FS Learning (Misc.)& 27.22 & \textbf{27.31} & 27.23 & 26.95 \\
    Synthetic (Misc.)  & 68.19 & 61.76 & 68.13 & \textbf{68.47}\\
    Code (Edit Sim)  & 25.02 & 24.68 & \textbf{25.58} & 24.21 \\
    \addlinespace[2pt]

    \multicolumn{5}{@{}l}{\textbf{RULER}}\\
    Retrieval (EM)     & \textbf{92.72} & 6.14 & 60.43 & 87.38 \\
    MH Tracing (EM)    & 68.08 & 3.64 & 65.24 & \textbf{68.48} \\
    Aggregation (EM)   & 36.07 & \textbf{37.42} & 24.92 & 27.74 \\
    QA (F1)            & 59.13 & 45.82 & 58.78 & \textbf{59.68} \\
    \addlinespace[2pt]

    \multicolumn{5}{@{}l}{\textbf{$\infty$-Bench}}\\
    En.QA (F1)         & \textbf{34.82} & 27.57& 34.69 & 32.44 \\
    Passkey (EM)       & 100.00 & 6.78 & 100.00 & 100.00 \\
    Number (EM)        & \textbf{99.32} & 6.44 & 96.10 & 92.20 \\
    KV (EM)            & \textbf{87.00} & 2.4 & 1.00  & 59.00 \\
    \bottomrule
  \end{tabularx}
\end{table}


We further evaluate the efficacy of SnapStream in a real deployment scenario with the mapping of DeepSeek-R1 described in Section~\ref{subsec:mapping} on a SN40L-16 node. We extend the SnapStream maximum KV cache length to 32k and evaluate the deployment on an improved long sequence benchmark, LongBench-v2, and two short-input reasoning tasks, AIME24 and LiveCodeBench. The results in Table~\ref{tab:deepseek_acc} show that this real-life SnapStream deployment is able to maintain accuracy with respect to the full KV cache baselines, and in particular, SnapStream is able to outperform a similar deployment of fixed 32k sequence length \textit{without} SnapStream on LongBench-v2.

\begin{table}[!t]
  \centering
  \caption{Benchmark results for DeepSeek-R1-0528 on long sequence and reasoning tasks. ``Full'' and ``32k'' refer to full attention baselines with maximum sequence lengths of 128k and 32k respectively. ``SnapStream-32k'' is run with a maximum sequence length of 128k and a fixed 32k KV cache length.}
  \setlength{\tabcolsep}{3.5pt}      
  \renewcommand{\arraystretch}{1.12} 
  \footnotesize                      
  \begin{tabularx}{\columnwidth}{l l c c c}
    \toprule
    \textbf{Benchmark} & \textbf{Metric} & Full & 32k & SnapStream-32k \\
    \midrule

    LongBench-v2 & Acc & \textbf{57.26} & 31.41 & 48.71 \\
    AIME24 & EM & \textbf{93.33} & 86.67 & 90.0 \\
    LiveCodeBench & pass@1 & 71.43 & 73.97 & \textbf{78.08} \\
    \bottomrule
  \end{tabularx}
   \label{tab:deepseek_acc}
\end{table}

\begin{table}[!t]
  \centering
  \caption{Mean input/output token lengths for accuracy benchmarks on DeepSeek-R1-0528. Completions are sampled with greedy decoding and a maximum sequence length of 128k.}
  \setlength{\tabcolsep}{3.5pt}      
  \renewcommand{\arraystretch}{1.12} 
  \footnotesize                      
  \begin{tabularx}{\columnwidth}{X X X}
    \toprule
    \textbf{Benchmark} & Input Length & Output Length \\
    \midrule

    LongBench-v2 & 84,762 & 1,296\\
    AIME24 & 140 & 15,424 \\
    LiveCodeBench & 476 & 10,634\\
    \bottomrule
  \end{tabularx}
   \label{tab:deepseek_tokens}
\end{table}

\subsection{Performance}

\begin{table}[t]
  \centering
  \small
  \caption{Normalized latency breakdown per kernel for one prefill MoE layer of DeepSeek with input sequence size of 128K.}
  \label{tab:prefill-latency-breakdown}
  \begin{tabularx}{\columnwidth}{l>{\raggedleft\arraybackslash}X}
    \toprule
    \textbf{Fused Kernels} & \textbf{Normalized Latency} \\
    \midrule
    QKV & 4.10\% \\
    MHA & 69.40\% \\
    Output Proj & 5.84\% \\
    Router & 3.15\% \\
    MoE FFN & 14.67\% \\
    SnapKV Compression & 2.52\% \\
    Streaming Ring Buffer & 0.32\% \\
    \bottomrule
  \end{tabularx}
\end{table}

\begin{table*}[!t]
  \centering
  \caption{Effect of SnapStream on maximum decode batch size and throughput for different prefill sequence sizes of a deployment of DeepSeek-R1-0528 on a SN40L-16 node.}
  \label{tab:snapstream-results}
  \begin{tabular}{ccccccc}
    \toprule
    \thead{Prefill\\Seq. Size} &
    \thead{Compressed\\Seq. Size} &
    \thead{Max Batch Size\\(No SnapStream)} &
    \thead{Max Batch Size\\(SnapStream)} &
    \thead{Throughput\\(No SnapStream)} &
    \thead{Throughput\\(SnapStream)} &
    \thead{Throughput\\Improvement} \\
    \midrule
    128K & 32K & 16  & 64  & 434  & 1832 & 4.2$\times$ \\
    64K  & 16K & 32  & 128 & 908  & 3928 & 4.3$\times$ \\
    32K  & 8K  & 64  & 256 & 1832 & 7903  & 4.3$\times$ \\
    \bottomrule
  \end{tabular}
\end{table*}

In this section, we evaluate the impact of SnapStream on memory usage, maximum batch size, and overall decoding throughput. All experiments are conducted with DeepSeek-R1-0528 under the three KV cache compression configurations shown in Table~\ref{tab:snapstream-results}. As described in Section~\ref{subsec:prefill}, SnapKV compression is applied during the prefill stage. The additional compression logic introduces a modest latency overhead of approximately 2--5\% of the total prefill time in the evaluated configurations. Table~\ref{tab:prefill-latency-breakdown} presents the normalized latency breakdown for one MoE layer during prefill for DeepSeek-R1 with input sequence size of 128k. As expected, we find that the MHA and MoE FFN kernels dominate TTFT, and that the two additional SnapStream kernels contribute to $<3\%$ of total latency.

SnapStream compression enables larger decoding batch sizes within the fixed memory budget of a 16-socket node, assuming the model runs entirely in HBM. For long sequence sizes, the KV-cache contributes significantly to the memory footprint; to avoid replicating the KV-cache across sockets, we run the MLA operator with $\text{DP16}$. Combined with the $4\times$ compressed KV-cache, this results in a $4\times$ increase in the maximum attainable batch size, as summarized in Table~\ref{tab:snapstream-results}.

Table~\ref{tab:snapstream-results} also shows the impact of KV-cache compression on decoding throughput when using the maximum batch size allowed for each sequence length. Throughput was measured on a single SN40L-16 node and averaged over 10 runs. The reported values are scaled by 2.4 to account for the average number of tokens accepted per forward pass, assuming an 80\% acceptance rate in Multi-Token Prediction (MTP). The combination of a smaller memory footprint and higher batch-level parallelism yields substantial end-to-end throughput gains — up to $4.3\times$ — for the evaluated configurations. These results highlight that SnapStream not only enables longer sequences and larger batches within the same memory budget but also directly improves decoding performance through better hardware utilization and reduced memory bandwidth pressure.

\section{Related Work}
Sparse attention has received sustained research interest since the introduction of the Transformer architecture~\cite{vaswani2023attentionneed}. Early works are primarily motivated by improving the $O(L^2)$ complexity of attention with sparse factorizations~\cite{child2019generatinglongsequencessparse}, fixed sparse patterns~\cite{beltagy2020longformerlongdocumenttransformer}, or locality-sensitive hashing~\cite{kitaev2020reformerefficienttransformer}. However, such methods require training from scratch with these modifications, leading to such methods falling out of favor as inroads to better model performance were made by simply scaling pretraining tokens and parameter counts. Recent SoTA models have found some success by applying sliding-window attention to a subset of their layers~\cite{gemmateam2024gemma2improvingopen,openai2025gptoss120bgptoss20bmodel}.

In response to the increased cost of LLM training, researchers have developed a variety of training-free KV cache compression methods. StreamingLLM~\cite{xiao2024efficientstreaminglanguagemodels} was originally developed as a method to enable LLMs to generate tokens beyond the maximum sequence length encountered during training, and allows for coherent text generation with a fixed KV cache budget, although accuracy may suffer. Quest~\cite{tang2024questqueryawaresparsityefficient} extends PagedAttention~\cite{kwon2023efficientmemorymanagementlarge} by offloading KV pages with low attention scores based on pre-computed statistics. InfLLM~\cite{xiao2024infllmtrainingfreelongcontextextrapolation} similarly offloads KV cache blocks to host memory and operates with a similar KV cache structure as SnapStream, but looks up Top-$K$ token blocks for every newly decoded token. SAGE-KV~\cite{wang2025llmsknowdropselfattention} and SnapKV~\cite{li2024snapkvllmknowslooking} perform one-time and length-thresholded KV cache compression respectively, but do not distinguish between prefill and decoding. SnapStream builds on training-free compression literature by combining the one-time compression of SnapKV with efficient decoding using a fixed budget from StreamingLLM, fitting each technique into the prefill-decode scheduling of a production LLM server. We leave incorporating pre-computed, offloaded KV blocks/pages in the style of Quest and InfLLM to future work.

However, training-free approaches invariably lead to some accuracy degradation, leading to a resurgence in trainable sparse attention. Cartridges~\cite{eyuboglu2025cartridgeslightweightgeneralpurposelong} propose compressing large corpuses into trainable KV caches with much smaller sequence lengths that may be optionally loaded using cached prefix lookups. Titans~\cite{behrouz2024titanslearningmemorizetest} propose to augment Transformers with a neural memory module that aims to memorize the data seen during training and improve efficiency and accuracy on long context tasks. DeepSeek-V2 introduced Multi-Head Latent Attention (MLA)~\cite{deepseekai2024deepseekv2strongeconomicalefficient}, which introduces shared low-rank projections of the query, key, and value vectors to reduce KV cache memory along the hidden dimension. However, doing so requires increased compute to project these compressed KVs back into the full hidden dimension during decoding. Native Sparse Attention (NSA)~\cite{yuan2025nativesparseattentionhardwarealigned} proposes a similar KV cache structure to SnapStream, with hierarchical compression of past tokens, that is trainable and aims to reduce complexity for both training and inference. DeepSeek Sparse Attention (DSA)~\cite{deepseekv32exp} proposes to finetune an existing dense attention model instead of training from scratch by training a token selection module, which still computes $O(L^2)$ scores between tokens, but with fewer heads and smaller hidden dimensions than full attention, thus reducing the compute costs of prefill. While SnapStream already accommodates MLA by compressing the smaller, low-rank latent vectors along the sequence dimension instead of the full-rank KVs, admitting the other architectural modifications mentioned above requires additional modifications to fused kernels that we leave to future work.

\section{Conclusion}
In this paper, we present SnapStream, a training-free KV cache compression method that employs SnapKV compression during prefill and a StreamingLLM rolling KV cache window during decoding. We describe the details of a static graph implementation of this method, as well as the details for a production 16-socket mixed tensor-parallel and data-parallel mapping of DeepSeek-R1-0528 on SN40Ls. This mapping uses SnapStream to reduce memory pressure and increase decoding throughput. We verify that SnapStream introduces minimal accuracy degradation on long sequence tasks for both Llama 3.1 8B Instruct and DeepSeek-R1, as well as additional reasoning and coding tasks for the latter model. Finally, we show how our SnapStream mapping enables us to achieve a $4.3\times$ improvement in decoding throughput, with at most a 5\% increase in prefill latency. We hope this work inspires further study of KV cache compression methods in production LLM serving workloads and a variety of hardware accelerators, and leave a closer examination of hyperparameters and mappings of other sparse attention mechanisms to future work.

\bibliographystyle{IEEEtran}
\bibliography{snapstream}

\clearpage
\appendix

\subsection{Experiment Hyperparameters}

In Table~\ref{tab:hyperparameters}, we show the SnapStream hyperparameters for Llama 3.1 8B Instruct experiments in Table~\ref{tab:llama_acc} and DeepSeek-R1-0528 experiments in Table~\ref{tab:deepseek_acc}. In Table~\ref{tab:hyper_ablations}, we ablate our hyperparemeter choices from those in Table~\ref{tab:hyperparameters} by varying the sink token length and top-$K$. We can see that when sink token length is reduced by $\frac{1}{4}$ to 128, LongBench-v2 performance drops by 11.73\%, where accuracy on other benchmarks remains relatively the same. Increasing the top-$K$ from 512 to 2048 doesn't appear to meaningfully change accuracy either; although AIME24 increases by 3\%, but that dataset has a very small sample size of 30 problems, so increasing solve rate by 1\% does not represent a meaningful change.

\begin{table}[t]
  \centering
  \caption{SnapStream Hyperparameters in Section~\ref{sec:experiments}}
  \resizebox{\columnwidth}{!}{%
\begin{tabular}{ccccccc}
    \toprule
    \thead{Model} &
    \thead{Sink} &
    \thead{Top-$K$} &
    \thead{Recent} &
    \thead{Total} &
    \thead{SnapKV \\Observation} &
    \thead{SnapKV \\Kernel} \\
    \midrule
    Llama & 768 & 6,400 & 1,024 & 8,192  & 32 & 13 \\
    DeepSeek & 512 & 512  & 31,744 & 32,768  & 32 & 13 \\
    \bottomrule
  \end{tabular}%
  }
  \label{tab:hyperparameters}
\end{table}

\begin{table}[!t]
  \centering
  \caption{SnapStream Hyperparameter ablations on DeepSeek-R1-0528. Baseline uses the hyperparameters in Table~\ref{tab:hyperparameters}.}
  \setlength{\tabcolsep}{3.5pt}      
  \renewcommand{\arraystretch}{1.12} 
  \footnotesize                      
  \begin{tabularx}{\columnwidth}{l l c c c}
    \toprule
    \textbf{Benchmark} & \textbf{Metric} & Baseline & Sink = 128 & Top-$K$ = 2048 \\
    \midrule

    LongBench-v2 & Acc & \textbf{48.71} & 36.98 & 48.31 \\
    AIME24 & EM & 90.0 & 90.0  & \textbf{93.3} \\
    LiveCodeBench & pass@1 & \textbf{78.08} & 76.32 & 76.32 \\
    \bottomrule
  \end{tabularx}
   \label{tab:hyper_ablations}
\end{table}

\subsection{SnapStream Static Graph Pseudo-code}
In this section, we show and describe PyTorch-style pseudocode for the operation of a SnapStream KV cache during prefill and decoding. We compare a na\"ive implementation of SnapStream with our ring buffer instantiation.

\subsubsection{Rolling Recent Window}
Listing~\ref{lst:naiveprefill} shows a na\"ive implementation of the SnapStream KV cache compression process with a rolling window buffer, similar to StreamingLLM. It uses the \verb!prefill_snapstream_topk! method described in Listing~\ref{lst:prefillsnapkv} to extract the top-$K$ evicted tokens between the sink and recent tokens. Construction of the compressed KV cache is then a straightforward concatenation in Lines 25-26.

\begin{lstlisting}[language=Python,float=*,caption=Na\"ive SnapStream Prefill KV Cache Compression,label=lst:naiveprefill]
def prefill_snapstream_naive_compression(key_states, value_states, attn_weights, sink_token_length, topk_length, recent_token_length, snapkv_observation_length, snapkv_kernel_size, prefill_length):
    '''
    key_states: (batch, n_head, max_seq_len, key_head_dim)
    value_states: (batch, n_head, max_seq_len, value_head_dim)
    attn_weights: (batch, n_head, max_seq_len, max_seq_len)
    sink_token_length: int
    topk_length: int
    recent_token_length: int
    snapkv_observation_length: int
    snapkv_kernel_size: int
    prefill_length: int
    '''
    batch, n_head, max_seq_len, key_head_dim = key_states.shape
    _, _, _, value_head_dim = value_states.shape
    sink_token_keys = key_states[:, :, :sink_token_length, :]
    sink_token_values = value_states[:, :, :sink_token_length, :]
    lhb = max(sink_token_length, prefill_length - recent_token_length)
    recent_token_keys = key_states[:, :, lhb:, :]
    recent_token_values = value_states[:, :, lhb:, :]
    topk_keys, topk_values = prefill_snapstream_topk(key_states, value_states, attn_weights, sink_token_length, topk_length, recent_token_length, snapkv_observation_length, snapkv_kernel_size, prefill_length)
    valid_topk = min(max(prefill_length - (sink_token_length + recent_token_length), 0), topk_length)
    compressed_keys = cat((sink_token_keys, recent_token_keys, topk_keys), dim=2)
    compressed_values = cat((sink_token_values, recent_token_values, topk_values), dim=2)
    return compressed_keys, compressed_values, valid_topk
    
\end{lstlisting}

Listing~\ref{lst:naivedecodeupdate} shows the corresponding na\"ive implementation of the SnapStream decoding phase KV cache update. In the case where \verb!cache_position! (sequence length $L$ in previous sections) is less than \verb!max_kv_length!, the update is a straightforward scatter (Lines 19-22). In the other case (Lines 24-32), we construct a new KV cache by concatenating slices of the sink, Top-$K$, and recent KVs. To accommodate different \verb!cache_position! values at different batch indices in a static graph, we perform both KV cache updates and select between the two depending on the batch-indexed cache position on Line 34. 

In practice, we find that the indexing and concatenation operations from Lines 24-31 in Listing~\ref{lst:naivedecodeupdate} cannot be parallelized along the sequence dimension and tend to take up significant amounts of on-chip memory. For a 32k sequence length KV cache, these non-parallelizable operations lead to the decoding stage described in Section~\ref{sec:decode-mapping} requiring 4.6$\times$ the amount of on-chip memory compared to the standard decoding kernel, with the indexing operations from Lines 24-29 by themselves taking up 323 MB of a single SN40L's 520MB of on-chip memory after sharding along the batch and head dimensions. In general, this implementation exacerbates the memory pressure of the decode section and makes it difficult to meet the overarching SLO of lower time per output token.


\begin{lstlisting}[language=Python,float=*,caption=Na\"ive SnapStream Decoding KV Cache Update,label=lst:naivedecodeupdate]
def decode_snapstream_naive(compressed_keys, compressed_values, decode_keys, decode_values, cache_position, sink_token_length, topk_length, recent_token_length, decode_length, valid_topk):
    '''
    compressed_keys: (batch, n_head, compressed_max_seq_len, key_head_dim) bf16
    compressed_values: (batch, n_head, compressed_max_seq_len, value_head_dim) bf16
    decode_keys: (batch, n_head, 1, key_head_dim) bf16
    decode_values: (batch, n_head, 1, value_head_dim) bf16
    cache_position: (batch, 1, 1) int32
    sink_token_length: int
    topk_length: int
    recent_token_length: int
    valid_topk: int
    '''
    max_kv_length = sink_token_length + recent_token_length + valid_topk
    # Branch 1: L < max_kv_length
    cache_position_bounded = where(cache_position >= max_kv_length, max_kv_length - 1, cache_position)
    branch1_compressed_keys = scatter(input=compressed_keys, dim=2, index=cache_position_bounded, src=decode_keys)
    branch2_compressed_values = scatter(input=compressed_values, dim=2, index=cache_position_bounded, src=decode_values)
    # Branch 2: L >= max_kv_length
    sink_keys = compressed_keys[:, :, :sink_token_length, :]
    sink_values = compressed_keys[:, :, :sink_token_length, :]
    topk_keys = compressed_keys[:, :, -valid_topk:, :]
    topk_values = compressed_values[:, :, -valid_topk:, :]
    new_recent_keys = compressed_keys[:, :, sink_token_length + 1:-topk_length]
    new_recent_values = compressed_values[:, :, sink_token_length + 1:-topk_length]
    branch2_compressed_keys = cat([sink_keys, new_recent_keys, decode_keys, topk_keys])
    branch2_compressed_values = cat([sink_values, new_recent_values, decode_values, topk_values])
    # select between branch 1 and branch 2
    selection_condition = cache_position.squeeze(-1, -2) < max_kv_length
    compressed_keys = where(selection_condition, branch1_compressed_keys, branch2_compressed_keys)
    compressed_values = where(selection_condition, branch1_compressed_values, branch2_compressed_values)
    return compressed_keys, compressed_values

    
\end{lstlisting}

\subsubsection{Ring Buffer}
To avoid these issues, we implement the KV cache as a ring buffer instead, as described in Section~\ref{subsec:prefill}. Listing~\ref{lst:prefillringbuffer} shows torch-style pseudocode implementing the construction of such a ring buffer during SnapStream prefill. 

\begin{lstlisting}[language=Python,float=*,caption=SnapStream Prefill Ring Buffer Construction,label=lst:prefillringbuffer]
def prefill_streamingllm_slicing(key_states, value_states, sink_token_length, recent_token_length, input_length):
    '''
    key_states: (batch, n_head, max_seq_len, key_head_dim)
    value_states: (batch, n_head, max_seq_len, value_head_dim)
    sink_token_length: int
    recent_token_length: int
    input_length: int
    '''
    batch, n_head, max_seq_len, key_head_dim = key_states.shape
    _, _, _, value_head_dim = value_states.shape
    target_seq_len = sink_token_length + recent_token_length
    streamllm_key_states = zeros(batch, n_head, target_seq_len, key_head_dim)
    streamllm_value_states = zeros(batch, n_head, target_seq_len, value_head_dim)
    sink_k = key_states[:, :, :sink_token_length, :]
    sink_v = value_states[:, :, :sink_token_length, :]
    # calculate left hand bound for ring buffer ranges
    range1_lhb_factor = (input_length - sink_token_length + recent_token_length) // recent_token_length
    range1_lhb = range1_lhb_factor * recent_token_length + sink - recent_token_length
    range1_lhb = max(range1_lhb, sink_token_length)
    range2_lhb = range1_lhb - recent_token_length
    range2_lhb = max(range2_lhb, sink_token_length)
    # selection between range 1 and range2 for ring buffer construction
    select_offset = input_length - range1_lhb
    range1_indices = arange(start=range1_lhb, end=range1_lhb + recent_token_length)[None, None, :]
    range2_indices = arange(start=range2_lhb, end=range2_lhb + recent_token_length)[None, None, :]
    range1_k = gather(key_states, range1_indices, dim=2)
    range2_k = gather(key_states, range2_indices, dim=2)
    range1_v = gather(value_states, range1_indices, dim=2)
    range2_v = gather(value_states, range2_indices, dim=2)
    ringbuffer_k = where(arange(recent_token_length)[None, None, :] <= select_offset, range1_k, range2_k)
    ringbuffer_v = where(arange(recent_token_length)[None, None, :] <= select_offset, range1_v, range2_v)
    streamingllm_k = cat(sink_k, ringbuffer_k)
    streamingllm_v = cat(sink_v, ringbuffer_v)
    return streamingllm_k, streamingllm_v
\end{lstlisting}

The code in Listing~\ref{lst:prefillringbuffer} is then used as a subroutine in Listing~\ref{lst:prefillcompression} to construct the ring buffer for the full SnapStream KV Cache during prefill. This greatly simplifies the decoding stage to a single scatter for each of the key and value tensors, as can be seen in Listing~\ref{lst:snapstreamdecodeupdate}.

\begin{lstlisting}[language=Python,float=*,caption=SnapStream Prefill KV Cache Compression,label=lst:prefillcompression]
def prefill_snapstream_compression(key_states, value_states, attn_weights, sink_token_length, topk_length, recent_token_length, snapkv_observation_length, snapkv_kernel_size, prefill_length):
    '''
    key_states: (batch, n_head, max_seq_len, key_head_dim)
    value_states: (batch, n_head, max_seq_len, value_head_dim)
    attn_weights: (batch, n_head, max_seq_len, max_seq_len)
    sink_token_length: int
    topk_length: int
    recent_token_length: int
    snapkv_observation_length: int
    snapkv_kernel_size: int
    prefill_length: int
    '''
    streamingllm_keys, streamingllm_values = prefill_streamingllm_slicing(key_states, value_states, sink_token_lnegth, recent_token_length, prefill_length)
    topk_keys, topk_values = prefill_snapstream_topk(key_states, value_states, attn_weights, sink_token_length, topk_length, recent_token_length, snapkv_observation_length, snapkv_kernel_size, prefill_length)
    valid_topk = min(max(prefill_length - (sink_token_length + recent_token_length), 0), topk_length)
    compressed_keys = cat((streamingllm_keys, topk_keys), dim=2)
    compressed_values = cat((streamingllm_values, topk_values), dim=2)
    return compressed_keys, compressed_values, valid_topk
    
\end{lstlisting}

\begin{lstlisting}[language=Python,float=*,caption=SnapStream Decoding KV Cache Update,label=lst:snapstreamdecodeupdate]
def decode_snapstream_naive(compressed_keys, compressed_values, decode_keys, decode_values, cache_position, sink_token_length, topk_length, recent_token_length, decode_length, valid_topk):
    '''
    compressed_keys: (batch, n_head, compressed_max_seq_len, key_head_dim) bf16
    compressed_values: (batch, n_head, compressed_max_seq_len, value_head_dim) bf16
    decode_keys: (batch, n_head, 1, key_head_dim) bf16
    decode_values: (batch, n_head, 1, value_head_dim) bf16
    cache_position: (batch, 1, 1) int32
    sink_token_length: int
    topk_length: int
    recent_token_length: int
    decode_length: (batch,) int32
    valid_topk: (batch,) int32
    '''
    max_kv_length = sink_token_length + recent_token_length + valid_topk
    ring_buffer_index = (decode_length - sink_token_length + recent_token_length) % recent_token_length + sink_token_length
    ring_buffer_index = where(decode_length < sink_token_length, decode_length, ring_buffer_index)
    compressed_keys = scatter(compressed_keys, index=cache_position, src=decode_keys, cache_position)
    compressed_values = scatter(compressed_values, index=cache_position, src=decode_values, cache_position)
    return compressed_keys, compressed_values

    
\end{lstlisting}

\begin{lstlisting}[language=Python,float=*,caption=SnapStream Prefill SnapKV Compression,label=lst:prefillsnapkv]
def prefill_snapstream_topk(key_states, value_states, attn_weights, sink_token_length, topk_length, recent_token_length, snapkv_observation_length, snapkv_kernel_size, prefill_length):
    '''
    key_states: (batch, n_head, max_seq_len, key_head_dim)
    value_states: (batch, n_head, max_seq_len, value_head_dim)
    attn_weights: (batch, n_head, max_seq_len, max_seq_len)
    sink_token_length: int
    topk_length: int
    recent_token_length: int
    snapkv_observation_length: int
    snapkv_kernel_size: int
    prefill_length: int
    '''
    attn_weights_sum = attn_weights[:, :, prefill_length - snapkv_observation_length, prefill_length, :prefill_length - snapkv_observation_length].sum(dim=2)
    snapkv_attn_weights = pool2d(attn_weights_sum, kernel_size=snapkv_kernel_size, stride=1, padding=snapkv_kernel_size//2)
    snapkv_attn_weights = snapkv_attn_weights[:, :, sink_token_length:-recent_token_length]
    topk_indices = topk(snapkv_attn_weights, topk_length, dim=-1) # (batch, n_head, topk_length)
    topk_keys = gather(key_states, sink_token_length + topk_indices, dim=2)
    topk_values = gather(value_states, sink_token_length + topk_indices, dim=2)
    return topk_keys, topk_values
    
    
\end{lstlisting}

\end{document}